# Quantum Edge Detection for Image Segmentation in Optical Environments

**Mario Mastriani**

DLQS LLC, 4431 NW 63RD Drive, Coconut Creek, FL 33073, USA.
mmastri@gmail.com

*Abstract*—A quantum edge detector for image segmentation in optical environments is presented in this work. A Boolean version of the same detector is presented too. The quantum version of the new edge detector works with computational basis states, exclusively**.** This way, we can easily avoid the problem of quantum measurement retrieving the result of applying the new detector on the image. Besides, a new criterion and logic based on projections onto vertical axis of Bloch's Sphere exclusively are presented too. This approach will allow us: 1) a simpler development of logic quantum operations, where they will closer to those used in the classical logic operations, 2) building simple and robust classical-to-quantum and quantum-to-classical interfaces. Said so far is extended to quantum algorithms outside image processing too. In a special section on metric and simulations, a new metric based on the comparison between the classical and quantum versions algorithms for edge detection of images is presented. Notable differences between the results of classical and quantum versions of such algorithms (outside and inside of quantum computer, respectively) show the existence of implementation problems involved in the experiment, and that they have not been properly modeled for optical environments. However, although they are different, the quantum results are equally valid. The latter is clearly seen in the computer simulations.

## 1 Introduction

Quantum computation and quantum information is the study of the information processing tasks that can be accomplished using quantum mechanical systems. Like many simple but profound ideas it was a long time before anybody thought of doing information processing using quantum mechanical systems [1].

Quantum computation is the field that investigates the computational power and other properties of computers based on quantum-mechanical principles. An important objective is to find quantum algorithms that are significantly faster than any classical algorithm solving the same problem. The field started in the early 1980s with suggestions for analog quantum computers by Paul Benioff [2] and Richard Feynman [3, 4], and reached more digital ground when in 1985 David Deutsch defined the universal quantum Turing machine [5]. The following years saw only sparse activity, notably the development of the first algorithms by Deutsch and Jozsa [6] and by Simon [7], and the development of quantum complexity theory by Bernstein and Vazirani [8]. However, interest in the field increased tremendously after Peter Shor's very surprising discovery of efficient quantum algorithms (or simulations on a quantum computer) for the problems of integer factorization and discrete logarithms in 1994 [9].

Since most of current classical cryptography is based on the assumption that these two problems are computationally hard, the ability to actually build and use a quantum computer would allow us to break most current classical cryptographic systems, notably the Rivest, Shamir y Adleman (RSA) system [10, 11]. In contrast, a quantum form of cryptography due to Bennett and Brassard [12] is unbreakable even for quantum computers.

On the other hand, and as well say Hirota *et al* inside the Introduction of their work [13]:

*Quantum computation has appeared in various areas of computer science such as information theory, cryptography, image processing, etc. [1] because there are inefficient tasks on classical computers that can be overcome by exploiting the power of the quantum computation. Processing and analysis of images in particular and visual information in general on classical computers have been studied extensively [14-17]. On quantum computers, the research on images has faced fundamental difficulties because the field is still in its infancy. To start with, what are quantum images or how do we represent images on quantum computers? Secondly, what should we do to prepare and process the quantum images on quantum computers?*

Precisely, these two questions represent the essence on which this paper is based, i.e., the correct (and more efficient) internal representation of an image in a quantum context, and its recovery, once processed internally. Thus, we recognize only 3 milestones in the brief history of quantum image processing, namely:

- all starts with the pioneering work of Prof. Salvador E. Venegas-Andraca [18-21] at Keble College, Oxford University, UK (currently at Tecnológico de Monterrey, Campus Estado de México), where he proposes quantum image representations such as Qubit Lattice [22], in fact, this is the first doctoral thesis in the specialty,

- the history continues with the quantum image representation via the Real Ket [23] of Prof. Jose I. Latorre Sentís, at Universitat de Barcelona, Spain, with a special interest in image compression in a quantum context, and finally,

- we arrive at the proposal of Prof. Kaoru Hirota *et al* [13] from Tokyo Institute of Technology, for a flexible representation of quantum images to provide a representation for images on quantum computers in the form of a normalized state which captures information about colors and their corresponding positions in the images.

These works marked the path and viability of quantum image processing, however, we believe that a new type of internal representation of images, which enable an easier representation of traditional algorithms of traditional Digital Image Processing in a quantum computer, as well as more easy and efficient recovery of images processed outside the quantum computer is imperative. Besides, we present a novel proposal to recovery quantum state to the output of a quantum algorithm after its measurement via a modified Kalman's Filter [24-28], and Recursive Least Squares (RLS) filter [29-31], too. This is the essence of this work, which is organized as follows:

The basic principles of Quantum Information Processing are outlined in Section 2. Implementation Problems in Quantum Image Processing are presented in Section 3. The new approach for internal image representation is outlined in Section 4. In Section 5, we present the development of Quantum-Boolean Image Processing concept. In Section 6, we show the proposed new interfaces classical-to-quantum and quantum-to-classical arising from the tools mentioned in the previous sections. In Section 7, we present both, a new classical and a new quantum edge detectors. In Section 8, we discuss the more appropriate metric for edge detection in a set of experimental results. Finally, Section 9 provides a conclusion and future works proposal of the paper.

## 2 Quantum Information Processing

In this section, we present the main concepts related to Quantum Information Processing, that is to say: qubit, Bloch's Sphere, Hilbert's Space, Schrödinger Equation, Unitary Operators, Quantum Circuits/Gates, and Quantum Algorithms.

2.1 Quantum bits (qubits) and Bloch's sphere

The bit is the fundamental concept of classical computation and classical information. Quantum computation and quantum information are built upon an analogous concept, the quantum bit, or qubit for short. In this section we introduce the properties of single and multiple qubits, comparing and contrasting their properties to those of classical bits [1].

The difference between bits and qubits is that a qubit can be in a state other than $|0\rangle$ or $|1\rangle$ [32, 33]. It is also possible to form linear combinations of states, often called superpositions:

$$|\psi\rangle = \alpha|0\rangle + \beta|1\rangle, \tag{1}$$

where $|\alpha|^2 + |\beta|^2 = 1$, with the states $|\alpha\rangle$ and $|\beta\rangle$ are understood as different polarization states of light. The numbers $\alpha$ and $\beta$ are complex numbers, although for many purposes not much is lost by thinking of them as real numbers. Put another way, the state of a qubit is a vector in a two-dimensional complex vector space. The special states $|0\rangle$ and $|1\rangle$ are known as Computational Basis States (CBS), and form an orthonormal basis for this vector space, being

$$|0\rangle = \begin{bmatrix} 1 \\ 0 \end{bmatrix}$$

and

$$|1\rangle = \begin{bmatrix} 0 \\ 1 \end{bmatrix}$$

One picture useful in thinking about qubits is the following geometric representation.

Because $|\alpha|^2 + |\beta|^2 = 1$, we may rewrite Equation (1) as

$$|\psi\rangle = e^{i\gamma}\left(\cos\frac{\theta}{2}|0\rangle + e^{i\phi}\sin\frac{\theta}{2}|1\rangle\right) = e^{i\gamma}\left(\cos\frac{\theta}{2}|0\rangle + (\cos\phi + i\sin\phi)\sin\frac{\theta}{2}|1\rangle\right) \tag{2}$$

where $0 \leq \theta \leq \pi$, $0 \leq \phi < 2\pi$. We can ignore the factor of $e^{i\gamma}$ out the front, because it has no observable effects [1], and for that reason we can effectively write

$$|\psi\rangle = \cos\frac{\theta}{2}|0\rangle + e^{i\phi}\sin\frac{\theta}{2}|1\rangle \tag{3}$$

The numbers $\theta$ and $\phi$ define a point on the unit three-dimensional sphere, as shown in Fig. 1.

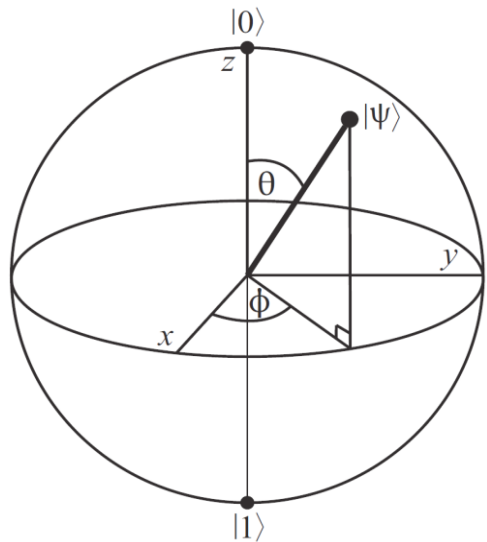

**Fig. 1** Bloch's Sphere.

Quantum mechanics is mathematically formulated in Hilbert space or projective Hilbert space. The space of pure states of a quantum system is given by the one-dimensional subspaces of the corresponding Hilbert space (or the "points" of the projective Hilbert space). In a two-dimensional Hilbert space this is simply the complex projective line, which is a geometrical sphere.

This sphere is often called the Bloch's sphere; it provides a useful means of visualizing the state of a single qubit, and often serves as an excellent testbed for ideas about quantum computation and quantum information. Many of the operations on single qubits which we describe later in this chapter are neatly described within the Bloch's sphere picture. However, it must be kept in mind that this intuition is limited because there is no simple generalization of the Bloch's sphere known for multiple qubits [1, 32, 33].

Except in the case where $|\psi\rangle$ is one of the ket vectors $|0\rangle$ or $|1\rangle$ the representation is unique. The parameters $\theta$ and $\phi$, re-interpreted as spherical coordinates, specify a point $\vec{a} = (sin\theta cos\phi + sin\theta sin\phi + cos\theta)$ on the unit sphere in $\mathbb{R}^3$ (according to Eq. 2).

Fig. 2 highlights all components (details) concerning the Bloch's sphere, namely

$$Spin\ down = |\downarrow\rangle = |0\rangle = \begin{bmatrix} 1 \\ 0 \end{bmatrix} = qubit\ basis\ state = North\ Pole$$

and

$$Spin\ up = |\uparrow\rangle = |1\rangle = \begin{bmatrix} 0 \\ 1 \end{bmatrix} = qubit\ basis\ state = South\ Pole$$

Both poles play a fundamental role in the development of the subsequent sections. Besides, a very important concept to the affections of the development of this work, i.e., the notion of latitude (parallel) on the Bloch's sphere is hinted. Such parallel as shown in green in Fig. 2, where we can see the complete coexistence of poles, parallels and meridians on the sphere, including computational basis states ($|0\rangle, |1\rangle$). The poles and the parallels form the geometric bases of criteria and logic needed to implement our algorithms for quantum image processing and the classical-to-quantum, and quantum-to-classical interfaces.

2.2 Schrödinger´s equation and unitary operators

A quantum state can be transformed into another state by a unitary operator, symbolized as $U$, with $U^\dagger U = I$ (where $I$ is the identity matrix), which is required to preserve inner products: If we transform $|\chi\rangle$ and $|\psi\rangle$ to $U|\chi\rangle$ and $U|\psi\rangle$, then $\langle\chi|UU|\psi\rangle = \langle\chi|\psi\rangle$. In particular, unitary operators preserve lengths: $\langle\psi|UU|\psi\rangle = \langle\psi|\psi\rangle = 1$.

On the other hand, the unitary operator satisfies the following differential equation known as the Schrödinger equation [1, 32-34]:

$$\frac{d}{dt}U(t) = \frac{-i\hat{H}}{\hbar}U(t) \qquad (4)$$

where $\hat{H}$ represents the Hamiltonian matrix of the Schrödinger equation, $i = \sqrt[2]{-1}$, and $\hbar$ is the Planck constant. Multiplying both sides of Eq.(4) by $|\psi(0)\rangle$ and setting $|\psi(t)\rangle = U(t)|\psi(0)\rangle$ yields

$$\frac{d}{dt}|\psi(t)\rangle = \frac{-i\hat{H}}{\hbar}|\psi(t)\rangle \qquad (5)$$

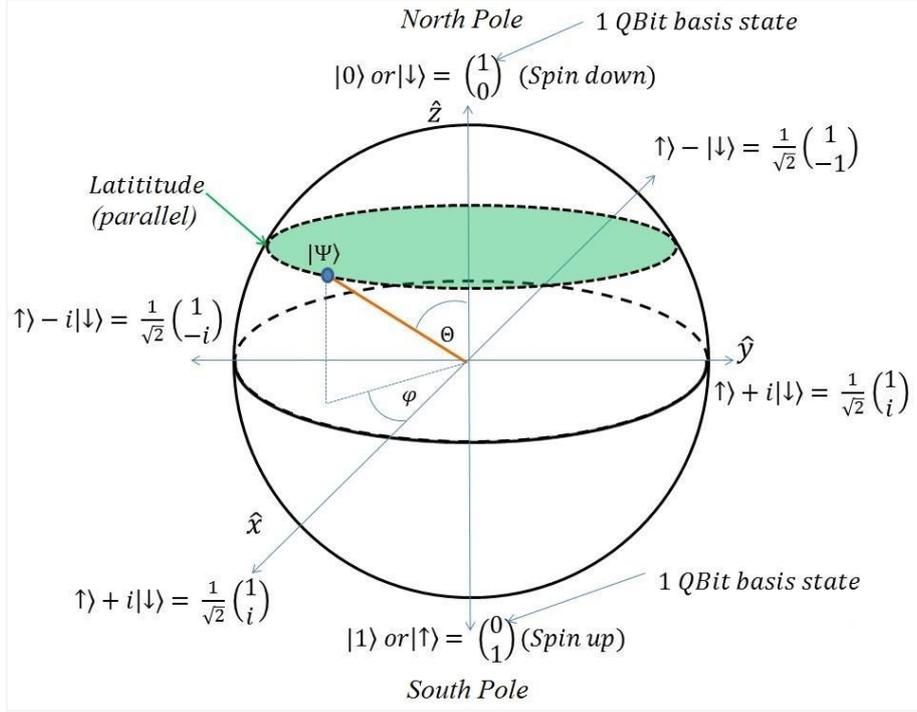

**Fig. 2** Details of the poles, as well as an example of parallel and several qubit states on the sphere.

The solution to the Schrödinger equation is given by the matrix exponential of the Hamiltonian matrix:

$$U(t) = e^{\frac{-i\hat{H}t}{\hbar}} \tag{6}$$

Thus the probability amplitudes evolve across time according to the following equation:

$$|\psi(t)\rangle = e^{\frac{-i\hat{H}t}{\hbar}} |\psi(0)\rangle \tag{7}$$

Equation 7 is the main piece in building circuits, gates and quantum algorithms, being $U$ who represents such elements [1].

Finally, the discrete version of Eq.(5) is

$$|\psi_{t+1}\rangle = \frac{-i\hat{H}}{\hbar} |\psi_t\rangle \tag{8}$$

Equation 8 is the foundation on which we build the optimal estimator of quantum states.

2.3 Quantum Circuits, Gates and Algorithms

As we can see in Fig. 3, and remember Eq.(8), the quantum algorithm (identical case to circuits and gates) viewed as a transfer (or mapping input-to-output) has two types on output:

*a)* the result of algorithm (circuit of gate), i.e., $|\psi_{t+1}\rangle$

b) part of the input $|\psi_t\rangle$, i.e., $|\underline{\psi}_t\rangle$ (underlined $|\psi_t\rangle$), in order to impart reversibility to the circuit, which is a critical need in quantum computing [1].

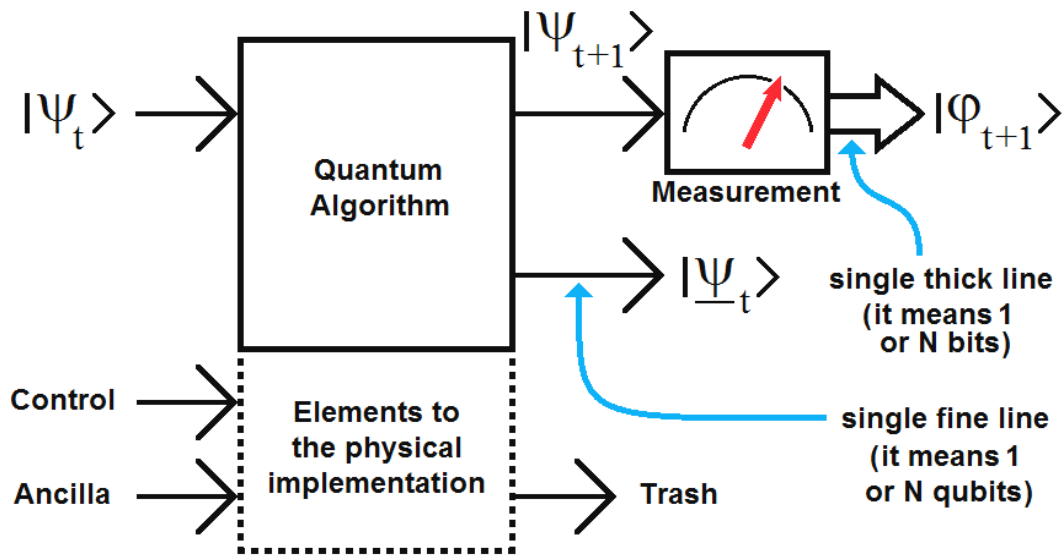

**Fig. 3** Module to measuring, quantum algorithm and the elements needs to its physical implementation.

Besides, we can see clearly a module for measuring $|\psi_{t+1}\rangle$ (which will be extensively discussed in the next section) with their respective output, i.e., $|\varphi_{t+1}\rangle$, and a number of elements needed for the physical implementation of the quantum algorithm (circuit or gate), namely: control, ancilla and trash [1]. In this figure as well as in the rest of them (unlike [1]) a single fine line represents a wire carrying *1* qubit or *N* qubits (qudit), interchangeably, while a single thick line represents a wire carrying *1* or *N* classical bits, interchangeably too.

However, the mentioned concept of reversibility is closely related to energy consumption, and hence to the Landauer's Principle [1].

On the other hand, computational complexity studies the amount of time and space required to solve a computational problem. Another important computational resource is energy. In [1], the authors shows the energy requirements for computation. Surprisingly, it turns out that computation, both classical and quantum, can in principle be done without expending any energy! Energy consumption in computation turns out to be deeply linked to the reversibility of the computation.

In other words, it is inexcusable the need of the $|\underline{\psi}_t\rangle$ presence to the output of quantum gate [1].

## 3 Implementation Problems in Quantum Image Processing

The implementation problems in Quantum Image Processing are:
- Wave function collapse,
- Quantum Measurement Problems,
- Before and after measurement,
- Types of measurement and state reconstruction,
- Interfaces, and
- Internal representations of an image and its possible implementations.

However, we present here the most important of them relatives to carry out quantum logic operations with CBS, which are fundamental concepts for the posterior development of our own internal representation of an image (inside quantum processor), classical-to-quantum and quantum-to-classical interfaces, and optimal state estimator after quantum measurement.

## 3.1 Before and after measurement

In quantum mechanics, measurement is a non-trivial and highly counter-intuitive process. Firstly, because measurement outcomes are inherently probabilistic, i.e. regardless of the carefulness in the preparation of a measurement procedure, the possible outcomes of such measurement will be distributed according to a certain probability distribution. Secondly, once a measurement has been performed, a quantum system in unavoidably altered due to the interaction with the measurement apparatus. Consequently, for an arbitrary quantum system, pre-measurement and post-measurement quantum states are different in general [22].

***Postulate.*** Quantum measurements are described by a set of measurement operators $\{\hat{M}_m\}$, index $m$ labels the different measurement outcomes, which act on the state space of the system being measured. Measurement outcomes correspond to values of *observables*, such as position, energy and momentum, which are Hermitian operators [1, 22] corresponding to physically measurable quantities.

Let $|\psi\rangle$ be the state of the quantum system immediately before the measurement. Then, the probability that result $m$ occurs is given by

$$p(m) = \langle\psi|\hat{M}_m^\dagger \hat{M}_m|\psi\rangle \tag{9}$$

and the post-measurement quantum state is

$$|\psi\rangle_{pm} = \frac{\hat{M}_m|\psi\rangle}{\sqrt{\langle\psi|\hat{M}_m^\dagger \hat{M}_m|\psi\rangle}} \tag{10}$$

Operators $\hat{M}_m$ must satisfy the completeness relation of Eq.(11a), because that guarantees that probabilities will sum to one; see Eq.(11b) [22]:

$$\sum_m \hat{M}_m^\dagger \hat{M}_m = I \tag{11a}$$
$$\sum_m \langle\psi|\hat{M}_m^\dagger \hat{M}_m|\psi\rangle = \sum_m p(m) = 1 \tag{11b}$$

Let us work out a simple example. Assume we have a polarized photon with associated polarization orientations 'horizontal' and 'vertical'. The horizontal polarization direction is denoted by $|0\rangle$ and the vertical polarization direction is denoted by $|1\rangle$. Thus, an arbitrary initial state for our photon can be described by the quantum state $|\psi\rangle = \alpha|0\rangle + \beta|1\rangle$, where $\alpha$ and $\beta$ are complex numbers constrained by the normalization condition $|\alpha|^2 + |\beta|^2 = 1$ and $\{|0\rangle, |1\rangle\}$ is the computational basis spanning $H^2$.

Now, we construct two measurement operators $\hat{M}_0 = |0\rangle\langle 0|$ and $\hat{M}_1 = |1\rangle\langle 1|$ and two measurement outcomes $a_0$, $a_1$. Then, the full observable used for measurement in this experiment is $\hat{M} = a_0|0\rangle\langle 0| + a_1|1\rangle\langle 1|$. According to Postulate, the probabilities of obtaining outcome $a_0$ or outcome $a_1$ are given by $p(a_0) = |\alpha|^2$ and $p(a_1) = |\beta|^2$. Corresponding post-measurement quantum states are as follows: if outcome = $a_0$ then $|\psi\rangle_{pm} = |0\rangle$; if outcome = $a_1$ then $|\psi\rangle_{pm} = |1\rangle$.

## 3.2 Interfaces

Figure 4 shows an overview of compression and decompression of a classic image thanks to the intervention of two quantum processors. The same scheme can be used in the context of filtering and image segmentation.

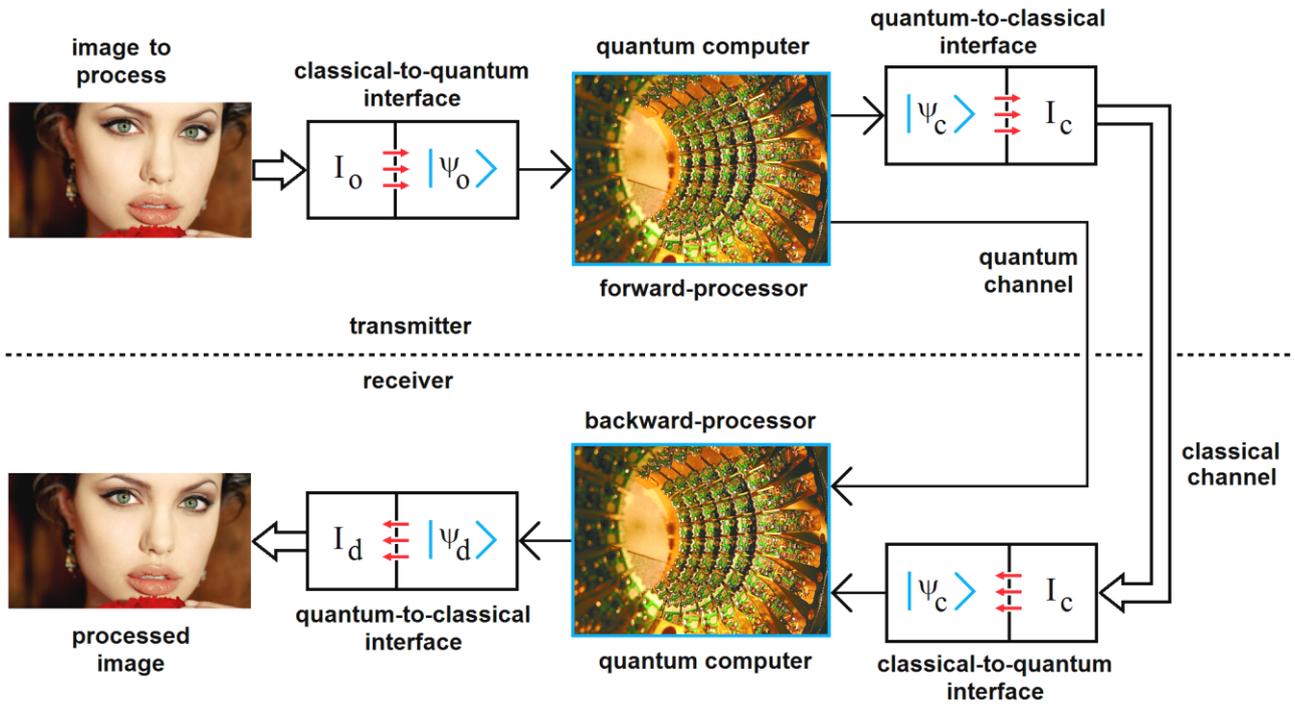

**Fig. 4** General scheme for quantum image processing.

For example, for the case of an image compression context, $I_o$ is the original image and $|\psi_o\rangle$ its counterpart (after classical-to-quantum interface), $|\psi_c\rangle$ is the quantum compression version of the image. If we use a quantum channel, $|\psi_c\rangle$ simply travels from transmitter to receiver. However, if we use a classical channel, it is imperative to use a quantum-to-channel interface, and $I_c$ is the traveling image.

On the other hand, inside receiver, $|\psi_c\rangle$ is decompressed and converted into $|\psi_d\rangle$. Other interface turns $|\psi_d\rangle$ on $I_d$, which is the decompressed classical image.

As we can see, it is absolutely inexcusable to use quantum-to-classical and classical-to-quantum interfaces, which to date is an unresolved problem.

3.3  Internal representations of an image and its possible implementations

Two problems hinder the development of the Quantum Image Processing, namely:
- The internal representation of a quantum image
- How introduce an image inside quantum computer and recover it after its later processing, i.e., the interfaces (classical-to-quantum, and quantum-to-classical, respectively).

Here we are going to mention the currently internal representations of an image inside a quantum machine, which were mentioned yet in Section 1:
- Qubit lattice [18-22],
- Real Ket [23], with a special interest in image compression in a quantum context, and finally,
- Flexible Representation of Quantum Images [13] to provide a representation for images on quantum computers in the form of a normalized state which captures information about colors and their corresponding positions in the images.

However, none of them has allowed (up to date) a practical implementation of any algorithm for quantum image processing. This is our challenge.

## 4 Pole-to-pole Axis Only (PAO)

According to Equations 1 and 3, as well as, Figures 1 and 2, $\alpha$ is the projection of $|\psi\rangle$ onto axis $z$, i.e.,

$$\alpha = \cos\frac{\theta}{2} \qquad (12)$$

As we can see in Fig. 5, for $|\psi_r\rangle$ and $|\psi_g\rangle$, $\alpha_r$ and $\alpha_g$ are their projections onto axis $z$, respectively. It is obvious that $|\psi_r\rangle$ has a greater latitude than $|\psi_g\rangle$, i.e., $\alpha_r > \alpha_g$. In other words, $\alpha_r$ is closer $|0\rangle$ (North Pole) than $\alpha_g$. Conversely, $\alpha_r$ is further $|1\rangle$ (South Pole) than $\alpha_g$.

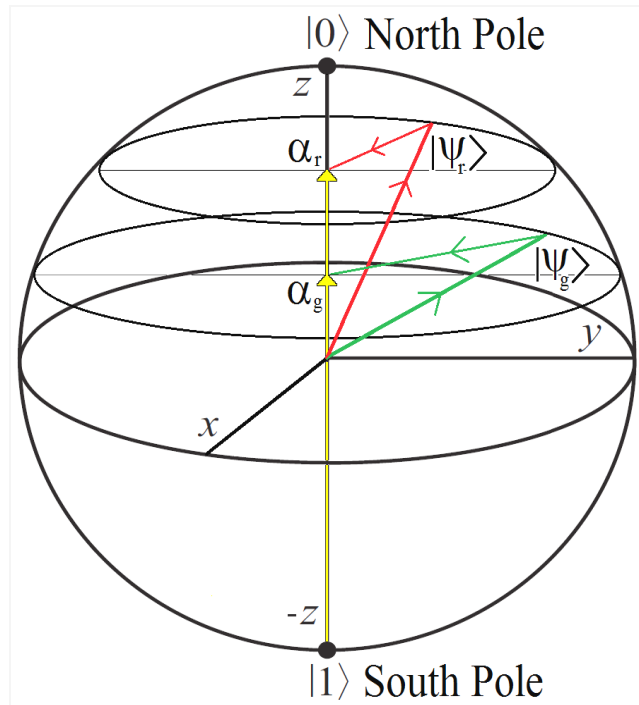

**Fig. 5** Projections onto $z$ axis.

Clearly, the projections on the $x$ and $y$ axis were completely obviated in our previous analysis. This is because such projections are absolutely irrelevant for practical purposes, constituting a principle called Pole-to-pole Axis Only (PAO), whereby, by which we can represent the pixels of an image (for each color) considering only values from the projections on the z axis.

Based on this simple fact, we can state new criteria, Logic and Arithmetic for the internal representation, processing and interface of images on a quantum computer.

4.1 PAO Criteria

Based on Fig. 6, if we do

$$\mu = 1 - \alpha \qquad (13)$$

we can see (for the same example) that $\mu_g > \mu_r$, that is to say, $\mu_r$ is closer $|0\rangle$ (North Pole) than $\mu_g$.

Conversely, $\mu_r$ is further $|1\rangle$ (South Pole) than $\mu_g$. Here seems to work with $\alpha$ and $\mu$ is the same, however, the difference is dramatic. To work with $\mu$ facilitate the internal representation of images, as well as classical-to-quantum and quantum-to-classical interface design and quantum algorithms processing.

Equation 13 receives the name of classical converter.

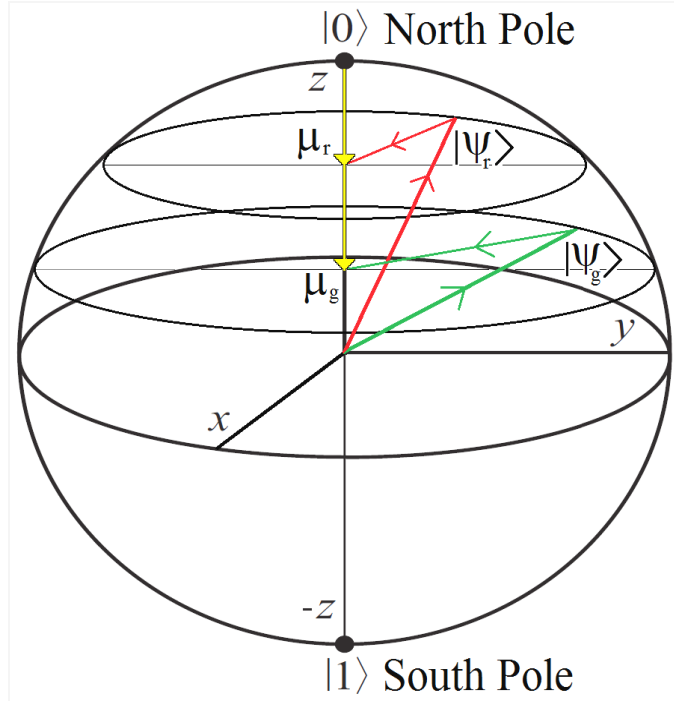

**Fig. 6** Converted projections onto *z* axis.

Its quantum counterpart is

$$|\mu\rangle = |0\rangle - |\psi\rangle \qquad (14)$$

and it is called quantum converter. Remember that $|0\rangle = \begin{bmatrix} 1 \\ 0 \end{bmatrix}$ and $|\psi\rangle = \begin{bmatrix} \alpha \\ \beta \end{bmatrix}$

then

$$\begin{bmatrix} \mu_\alpha \\ \mu_\beta \end{bmatrix} = \begin{bmatrix} 1 \\ 0 \end{bmatrix} - \begin{bmatrix} \alpha \\ \beta \end{bmatrix} = \begin{bmatrix} 1-\alpha \\ -\beta \end{bmatrix} \qquad (15)$$

In this approach the second component is absolutely overlooked, it has the value that it has, i.e., $-\beta$ or any other value, we do not care. Therefore, Equations 13 and 15 finally coincide. Thus, the POA criterion is:

*It only matters the projection onto z axis, that is, pole-to-pole axis, and their conversions.*

In case of several qubits, for example, in its generic form, we say B qubits, where B = 8 (in most of cases, concerning digital image processing inside quantum processor, and for each color [14-17]). Thus, in the last case, quantum numbers are

$$|00000000\rangle = \begin{pmatrix} 1 \\ 0 \\ \vdots \\ 0 \end{pmatrix} \Big\} 256\, elements \qquad (16)$$

$$|11111111\rangle = \begin{pmatrix} 0 \\ 0 \\ \vdots \\ 1 \end{pmatrix} \Big\} 256\, elements \qquad (17)$$

In both Equations, (16) and (17), each column has 1 *one* and 255 *zeros*. Otherwise, these numbers can be represented as $|0\rangle$ and $|2^B - 1\rangle = |255\rangle$ (with B = 8), respectively. In such case, $\alpha$ limits will be $min(\alpha) = 0$ and $max(\alpha) = 2^B - 1$, i.e., $0 \leq \alpha \leq 2^B - 1$. Therefore, classic converter is

$$\mu = (2^B - 1) - \alpha \qquad (18)$$

while the quantum converter is

$$|\mu\rangle = |0\rangle - |\psi\rangle \qquad (19)$$

That is to say, the same of Equation (14), however, $|0\rangle$ is associated with 256 levels instead of 2.

4.2 Logic operations according to PAO Criteria

For computational basis states ($|0\rangle, |1\rangle$), we can set four basic logic operations (AND, OR, XOR, and NOT), where $\overline{(\cdot)}$ means $NOT(\cdot)$, see Table I.

TABLE I
LOGIC OPERATIONS FOR COMPUTATIONAL BASIS STATES

| $|\psi_1\rangle$ | $|\psi_2\rangle$ | AND | OR | $\overline{|\psi_1\rangle}$ | $\overline{|\psi_2\rangle}$ | XOR |
|---|---|---|---|---|---|---|
| $|0\rangle$ | $|0\rangle$ | $|0\rangle$ | $|0\rangle$ | $|1\rangle$ | $|1\rangle$ | $|0\rangle$ |
| $|0\rangle$ | $|1\rangle$ | $|0\rangle$ | $|1\rangle$ | $|1\rangle$ | $|0\rangle$ | $|1\rangle$ |
| $|1\rangle$ | $|0\rangle$ | $|0\rangle$ | $|1\rangle$ | $|0\rangle$ | $|1\rangle$ | $|1\rangle$ |
| $|1\rangle$ | $|1\rangle$ | $|1\rangle$ | $|1\rangle$ | $|0\rangle$ | $|0\rangle$ | $|0\rangle$ |

being $|\psi_i\rangle = \begin{pmatrix} \alpha_i \\ \beta_i \end{pmatrix}$

Therefore,

$$\overline{|\psi_i\rangle} = NOT(|\psi_i\rangle) = NOT\begin{pmatrix} \alpha_i \\ \beta_i \end{pmatrix} = \begin{pmatrix} \beta_i \\ \alpha_i \end{pmatrix} \quad \forall i \qquad (20)$$

with

$$XOR(|\psi_1\rangle,|\psi_2\rangle) = (|\psi_1\rangle\, AND\, \overline{|\psi_2\rangle})\, OR\, (\overline{|\psi_1\rangle}\, AND\, |\psi_2\rangle) \qquad (21)$$

In traditional Quantum Logic, these four operations (AND, OR, XOR, and NOT) are implemented using the Control-NOT (CNOT) and Toffoli gates [1]. On the other hand, we can see what happens with $\alpha$ and $\mu$ in these cases.. In Table II, we have,

TABLE II
$\alpha$ AND $\mu$ FOR COMPUTATIONAL BASIS STATES

| $\alpha_1$ | $\alpha_2$ | $\mu_1$ | $\mu_2$ | AND | OR | XOR |
|---|---|---|---|---|---|---|
| 1 | 1 | 0 | 0 | $|0\rangle$ | $|0\rangle$ | $|0\rangle$ |
| 1 | 0 | 0 | 1 | $|0\rangle$ | $|1\rangle$ | $|1\rangle$ |
| 0 | 1 | 1 | 0 | $|0\rangle$ | $|1\rangle$ | $|1\rangle$ |
| 0 | 0 | 1 | 1 | $|1\rangle$ | $|1\rangle$ | $|0\rangle$ |

However, regarding to $\mu$, logics operations seems Boolean operations, see Table III. In fact, as we can see in Fig. 7, when result of AND operation is $|0\rangle$, one of the two $\mu$ equals to 0 and we are on North Pole, while in the case where AND operation is $|1\rangle$, for both $\mu$ equal to 1 and we are on South Pole.

TABLE III
LOGIC OPERATIONS REGARDING $\mu$

| $\mu_1$ | $\mu_2$ | $AND_\mu$ | $OR_\mu$ | $XOR_\mu$ |
|---|---|---|---|---|
| 0 | 0 | 0 | 0 | 0 |
| 0 | 1 | 0 | 1 | 1 |
| 1 | 0 | 0 | 1 | 1 |
| 1 | 1 | 1 | 1 | 0 |

Conversely, when result of OR operation is $|0\rangle$, for both $\mu$ equal to 0 and we are on North Pole, while in the case where OR operation is $|1\rangle$, with only one of them is 1. That is to say, inside PAO Criteria, AND operation is a minimum, while OR operation is a maximum between both $|\psi\rangle$. This is extended beyond the pure case of computational basis states ($|0\rangle$, $|1\rangle$). Then, we obtain

$$\begin{aligned}|\psi_1\rangle \wedge |\psi_2\rangle &= min(|\psi_1\rangle, |\psi_2\rangle) \\ |\psi_1\rangle \vee |\psi_2\rangle &= max(|\psi_1\rangle, |\psi_2\rangle) \\ |\psi_1\rangle \veebar |\psi_2\rangle &= max\left(min(|\psi_1\rangle, \overline{|\psi_2\rangle}), min(\overline{|\psi_1\rangle}, |\psi_2\rangle)\right)\end{aligned} \qquad (22)$$

where, $\wedge$ means AND, $\vee$ means OR, and $\veebar$ means XOR.

Here we can draw some conclusions:
- these logical operations for qubits placed serve anywhere in the Bloch's sphere (even if not pure qubits), thing that traditional quantum logic gates can not do [1, 35, 36].

- consistent with PAO criteria, it only imports projections on the vertical axis (pole-to-pole), i.e., it only care to know which is the highest or lowest parallel, in other words, the northernmost and southernmost parallel. Summing-up, we will consider only the projections on the vertical axis in the measurement process.

- these AND and OR logic operations give the same results as those obtained by the same operations in Fuzzy Logic [37-43]. i.e., *min*(.) and *max*(.), respectively. This allows us to imagine future applications of this technology applying it to Automatic Control [44, 45].

- then, this new logic is only possible after an exact measurement (without change on state, or wave collapse). If such thing is possible (thank to the estimator to be presented in this work) we can compare and order quantum states, which is impossible today [46, 47].

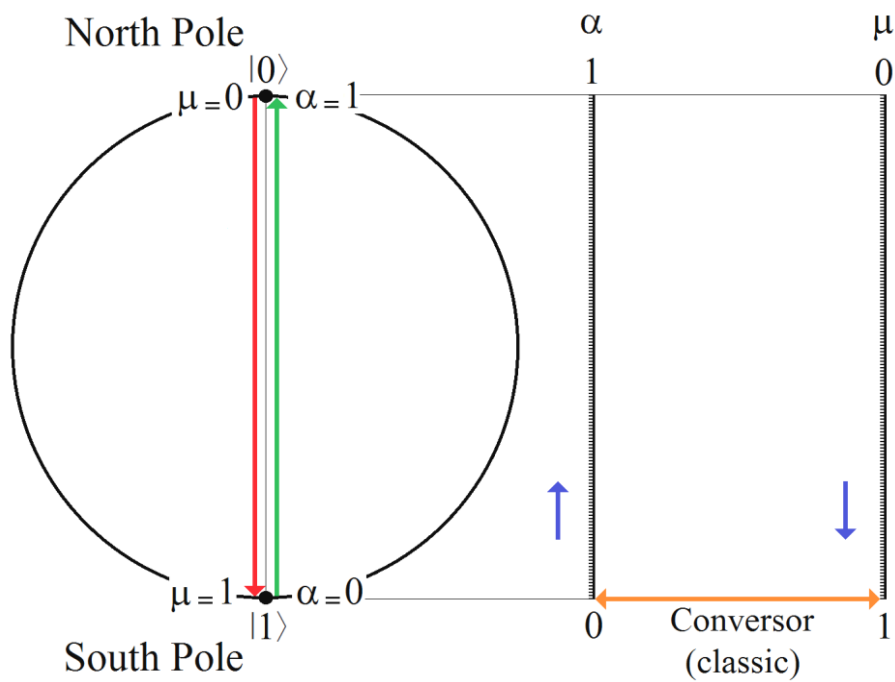

**Fig. 7** $\alpha$, $\mu$ and logic operation results on Bloch's sphere.

This new logic is better even than Multivalued Quantum Logic [48].

*For B qubits (qudit)*

The results are similar in shape to the previous case. Two things changed:

- we replace 1 by $2^B-1$, i.e., a quantum word called *quantum byte* or qubyte.
- the size of the qubits involved is substantially higher. However, here also, it only interest us the projection onto the pole-to-pole axis of the result vector (PAO criteria).

Fig. 7 show us the relationship between $\alpha$ and $\mu$, specifically, two examples, the red one ($\alpha_r$, $\mu_r$), and the green one ($\alpha_g$, $\mu_g$). If the final processed image *I* is Boolean (i.e, bi-level), then, the relationship between $\mu$ and *I* is direct. In return, if the final processed image *I* is discrete (i.e., multilevel), then, it is necessary an equalizer and rounder between them [49].

Summing-up: Three important details to note are:
- the progress of *I* coincides with the progress of $\mu$, fully backward to $\alpha$.
- the scales of $\mu$ and hence *I* are nonlinear
- it is necessary an equalizer and rounder between $\mu$ and hence *I* in nonbinary cases only.

## 5  Quantum-Boolean Image Processing (QuBoIP)

QuBoIP begin with an image conversion from color to gray version, as we can see on Fig.8. A possibility for this conversion is thanks to *rgb2gray(.)* built-in MATLAB® function [50].

Then, QuBoIP continues with a slicing of the image, in this case thanks to an own MATLAB® function called *slicer(.)*, and from which we get many bitplanes as depth in bit has the image to be treated. In Fig.8, we get 8 bitplanes, where, bitplane 7 is called Most Significant Bit (MSB) and it is the most morphologically committed bitplane with the gray image [51]. In return, bitplane 0 is the Least Significant Bit (LSB) and it is the least morphologically committed bitplane with the gray image.

Two important aspects:
- From here to the end of this paper, we are going to work with MSB, i.e., with we will say "image", we are saying MSB.
- The classical version of the *slicer(.)* function in MATLAB® code is:

```
function Ibpp = slicer(I,bpp)              function vp = d2b(p,bpp)

% NOTE:                                    % NOTE:
% bpp = bit-per-pixel                      % d = bit depth
% I = Image                                % p = pixel value
% Ibpp = Image in bpp bitplanes            % vp = binary vector for pixel

[ROW,COL] = size(I);                       vp = zeros(1,bpp);
for r = 1:ROW                              d = 1;
  for c = 1:COL                            while p > 0,
    aux = d2b(I(r,c)-1,bpp);                 vp(d) = mod(p,2);
    for b = 1:bpp                            p = p/2;
      Ibpp(r,c,b) = aux(b);                  p = floor(p);
    end                                      d = d+1;
  end                                      end
end                                        vp = rot90(rot90(vp));

return;                                    return;
```

If we were to highlight the advantage of working in QuBoIP rather than in QuIP [13, 18-23], it would certainly be the fact that as QuBoIP working with CBS exclusively, the measurement is not a problem as in the rest of Quantum Physics, because, when we measured an $|1\rangle$, the result is an unchanged $|1\rangle$, and when we measured an $|0\rangle$, the result is an unchanged $|0\rangle$.

Figure 9 show us -in detail- the 8 bitplanes of Angelina, from MSB (bitplane 7) to LSB (bitplane 0). Let observe that as we move from MSB to LSB, different bitplanes are increasingly unrecognizable compared to the original image, i.e., Angelina. As we can see, LSB is completely different regarding to original Angelina morphology. This is one reason why the LSB is Steganography territory [51]. The other reason is that any change in the LSB does not produce visually detectable changes in the original image.

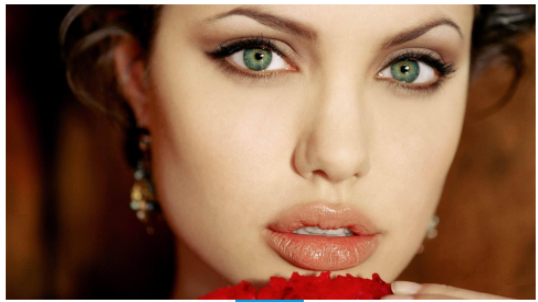

color version of Angelina

*rgb2gray(.)*

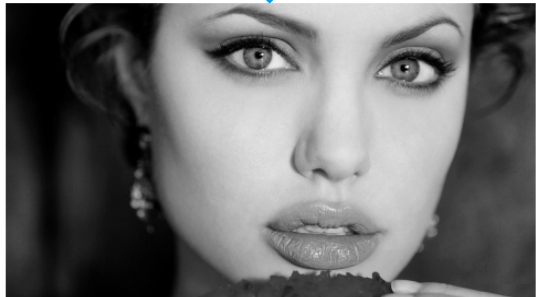

gray version of Angelina

*slicer(.)*

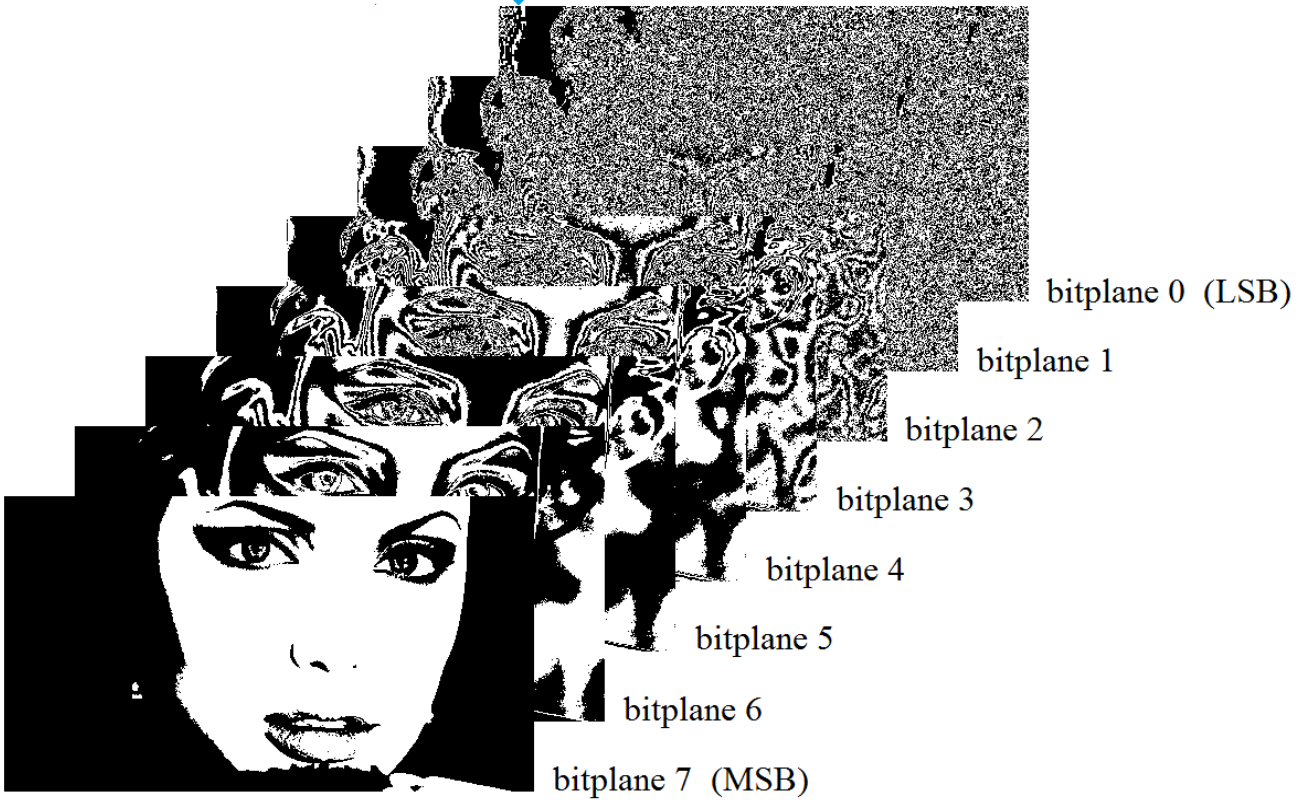

bitplane 0 (LSB)
bitplane 1
bitplane 2
bitplane 3
bitplane 4
bitplane 5
bitplane 6
bitplane 7 (MSB)

**Fig. 8** Color to gray conversion, slicing and obtaining of bitplanes, and MSB and LSB for Angelina.

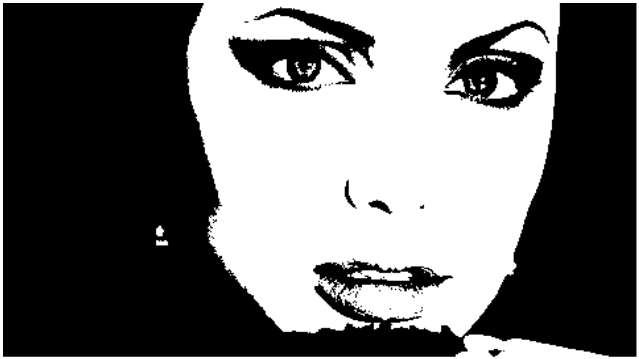
(MSB) bitplane 7

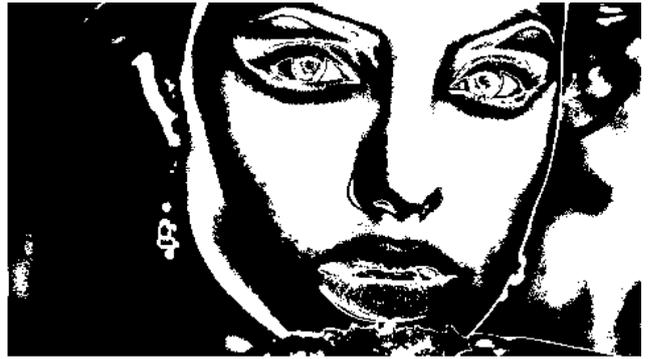
bitplane 6

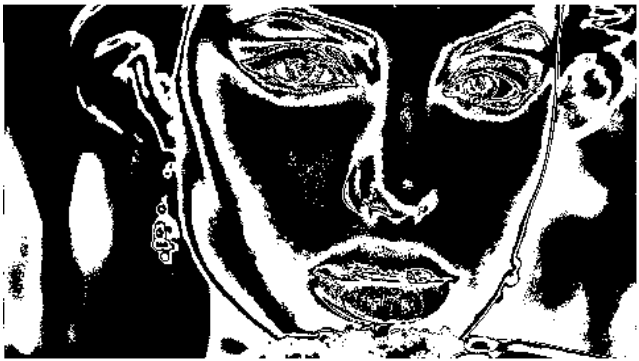
bitplane 5

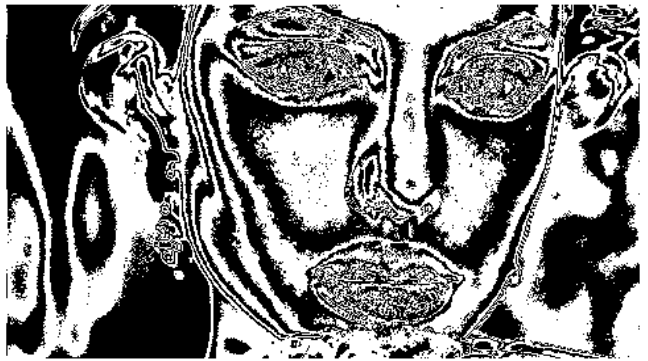
bitplane 4

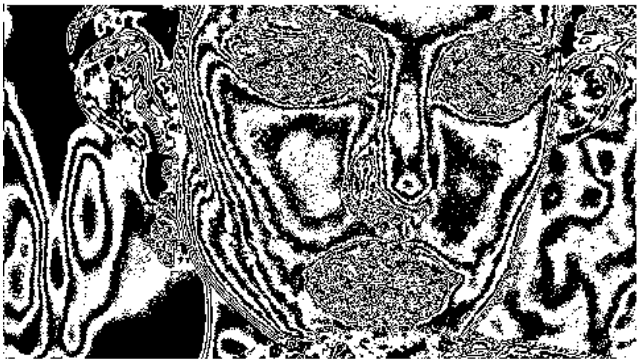
bitplane 3

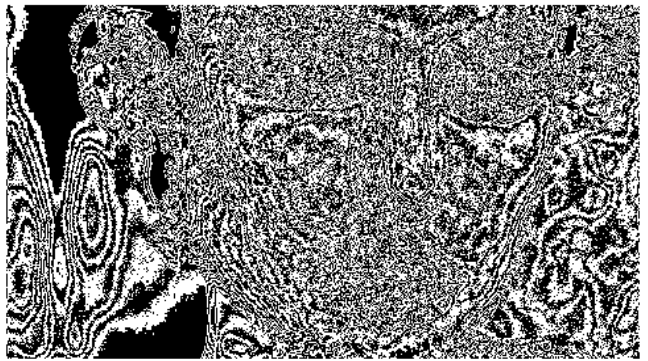
bitplane 2

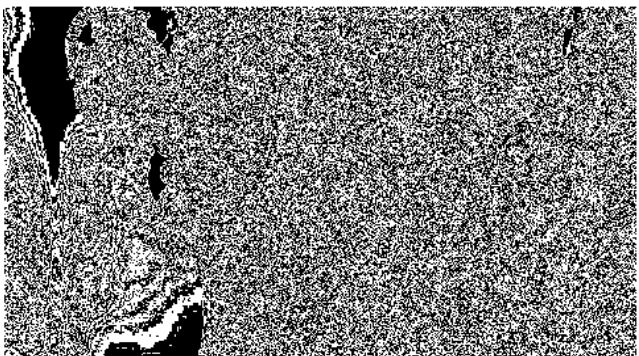
bitplane 1

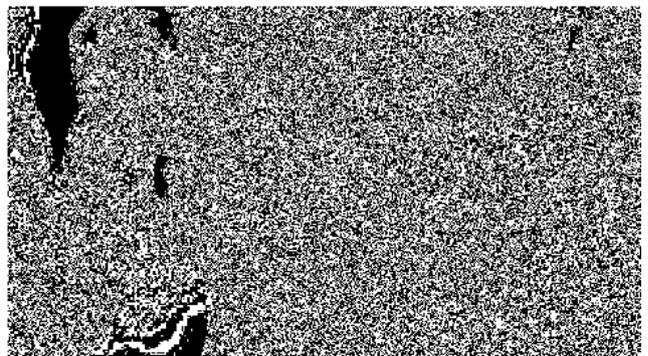
(LSB) bitplane 0

**Fig. 9** Angelina and her 8 bitplanes, including MSB and LSB.

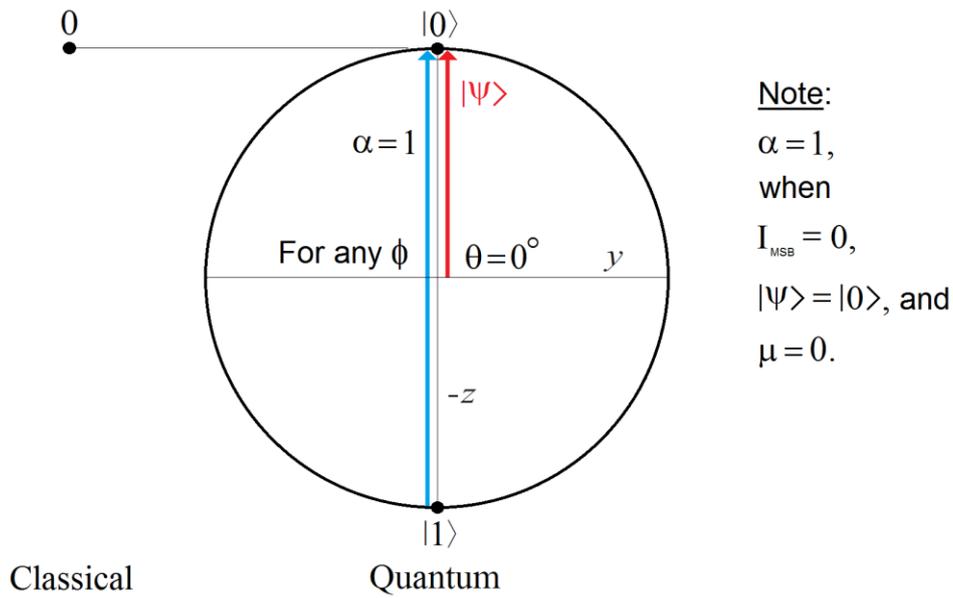

**Fig. 10** Relationship between classical 0, α, μ and $I_{MSB}$.

Figure 10 show as the relationship between classical 0, α, μ and $I_{MSB}$. In this case, α = 1 when $I_{MSB}$ = 0, $|\psi\rangle = |0\rangle$, and μ = 0, with θ = 0° and for any ϕ.

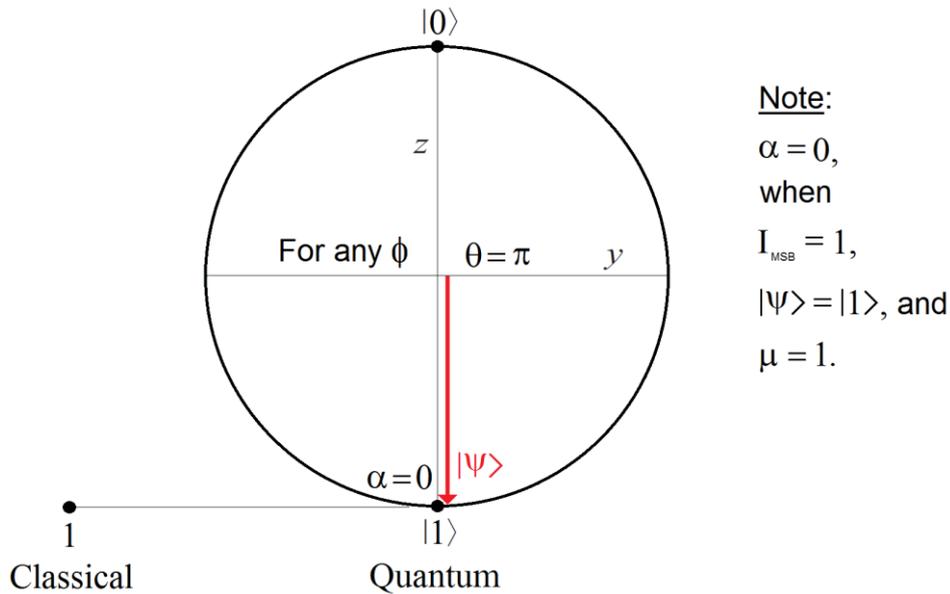

**Fig. 11** Relationship between classical 1, α, μ and $I_{MSB}$.

On the other hand, in Fig.11, we can see the relationship between classical 1, α, μ and $I_{MSB}$. This case is the opposite of the previous, with α = 0 when $I_{MSB}$ = 1, $|\psi\rangle = |1\rangle$, and μ = 1, with θ = π and for any ϕ too.

As we can see, the geometric relationship between Figures 10 and 11 is inverted. This happens to limiting values on the Bloch's sphere such as the CBS, that is to say, $|0\rangle$ and $|1\rangle$.

In the next sections, we are going to use these concepts for the purpose of enabling the construction of the best interfaces for our purposes, i.e., those required in QuBoIP.

# 6 Proposed Interfaces

Building interfaces between the classical and quantum world is an aspiration within this field [52-58]. Major obstacles in the construction of such interfaces are:
- decoherence [22, 59-64]
- functional interpretation of such an interface should work. This is the greatest contribution of PAO.

*What is quantum decoherence?*

In quantum mechanics, quantum decoherence is the loss of coherence or ordering of the phase angles between the components of a system in a quantum superposition. One consequence of this dephasing is classical or probabilistically additive behavior. Quantum decoherence gives the appearance of wave function collapse (the reduction of the physical possibilities into a single possibility as seen by an observer) and justifies the framework and intuition of classical physics as an acceptable approximation: decoherence is the mechanism by which the classical limit emerges from a quantum starting point and it determines the location of the quantum-classical boundary. Decoherence occurs when a system interacts with its environment in a thermodynamically irreversible way. This prevents different elements in the quantum superposition of the total scene's wavefunction from interfering with each other. Decoherence has been a subject of active research since the 1980s.

Decoherence can be viewed as the loss of information from a system into the environment (often modeled as a heat bath), since every system is loosely coupled with the energetic state of its surroundings. Viewed in isolation, the system's dynamics are non-unitary (although the combined system plus environment evolves in a unitary fashion). Thus the dynamics of the system alone are irreversible. As with any coupling, entanglements are generated between the system and environment. These have the effect of sharing quantum information with—or transferring it to—the surroundings.

Decoherence does not generate actual wave function collapse. It only provides an explanation for the observation of wave function collapse, as the quantum nature of the system "leaks" into the environment. That is, components of the wavefunction are decoupled from a coherent system, and acquire phases from their immediate surroundings. A total superposition of the global or universal wavefunction still exists (and remains coherent at the global level), but its ultimate fate remains as an interpretational issue. Specifically, decoherence does not attempt to explain the measurement problem. Rather, decoherence provides an explanation for the transition of the system to a mixture of states that seem to correspond to those states observers perceive. Moreover, our observation tells us that this mixture looks like a proper quantum ensemble in a measurement situation, as we observe that measurements lead to the "realization" of precisely one state in the "ensemble".

Decoherence represents a challenge for the practical realization of quantum computers, since such machines are expected to rely heavily on the undisturbed evolution of quantum coherences. Simply put, they require to preserve coherent states and to manage decoherence, in order to actually perform quantum computation. In short, the quantum decoherence is a trivial decoherence and unwanted passage between the quantum and the classical world.

*What is functional interpretation of such an interface should work?*

The answer to this question lies in the possibility offered PAO as a functional instrument that conveys building these interfaces. In this section, we developed two modes of quantum-to-classical interface and one mode of classical-to-quantum interface according to the above.

As we can see in previous sections, there is a direct and automatic correspondence between [0, 1] and [$|0\rangle$, $|1\rangle$]. Such correspondence (and in that order) will be the classis-to-quantum interface. In the same way, but in reverse order, there is a direct and automatic correspondence between [$|0\rangle$, $|1\rangle$] and [0, 1]. In this correspondence, but in that order, we know it as a quantum-to-classical interface. Unlike [49], the measurement is

TABLE IV
MEASUREMENT OUTCOME WITH CBS AND GENERIC STATE.

| Before quantum measurement | After quantum measurement |
|---|---|
| $\|0\rangle$ | $\|0\rangle$ |
| $\|1\rangle$ | $\|1\rangle$ |
| $\|\psi\rangle$ | $\|\psi\rangle_{pm} = \dfrac{\hat{M}_m \|\psi\rangle}{\sqrt{\langle\psi\|\hat{M}_m^\dagger \hat{M}_m\|\psi\rangle}}$ |

not a problem, since it does not alter the outcome measure. Therefore, it is not necessary and estimator after measurement as in [49]. In Table IV we can see these statements, where left column represents the state before measurement, while right column represents the state after that, for CBS and generic state (Eq.10).

6.1 Quantum-to-classical interface (Mode I)

In Fig. 12, we can see first the quantum algorithm, whose output is directed to the interface, which begins with the measurement operator, continuous with the OSE, and thanks to the first mode of PAO recovered $\mu_{\hat{\psi}}$, from $\alpha_{\hat{\psi}}$, so we use a classical converter, with $\mu_{\hat{\psi}} = (2^B - 1) - \alpha_{\hat{\psi}}$ for B qubits (here B = 1). Finally, we employed an equalizer and a rounder (which is not in the figure) to complete the scheme. This architecture is extensible to any dimension of qudits.

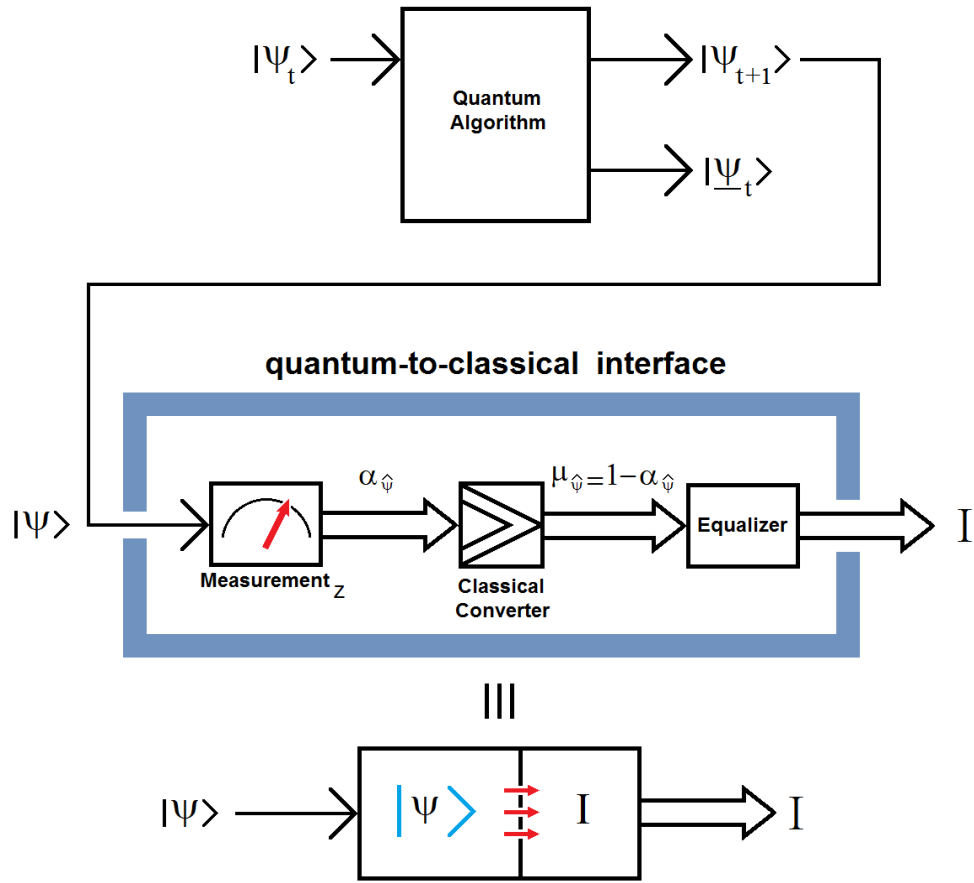

**Fig. 12** Quantum-to-classical interface (Mode I).

## 6.2 Quantum-to-classical interface (Mode II)

In Fig. 13, we can see first a quantum converter, then the quantum algorithm, whose output is directed to the interface, which begins with the measurement operator, continuous with the OSE, and thanks to the second mode of PAO recovered directly $\mu$, of which we are only interested in the projection on the vertical axis, i.e., $\alpha$, so we use the switch. Finally, we employed an equalizer and a rounder to complete the scheme [49]. This architecture is extensible to any dimension of qudits. The $|0\rangle$ of quantum converter is for $2^B = 2$ levels, i.e., B = 1 qubits. In other words, everything said here is for qubits and qudits interchangeably.

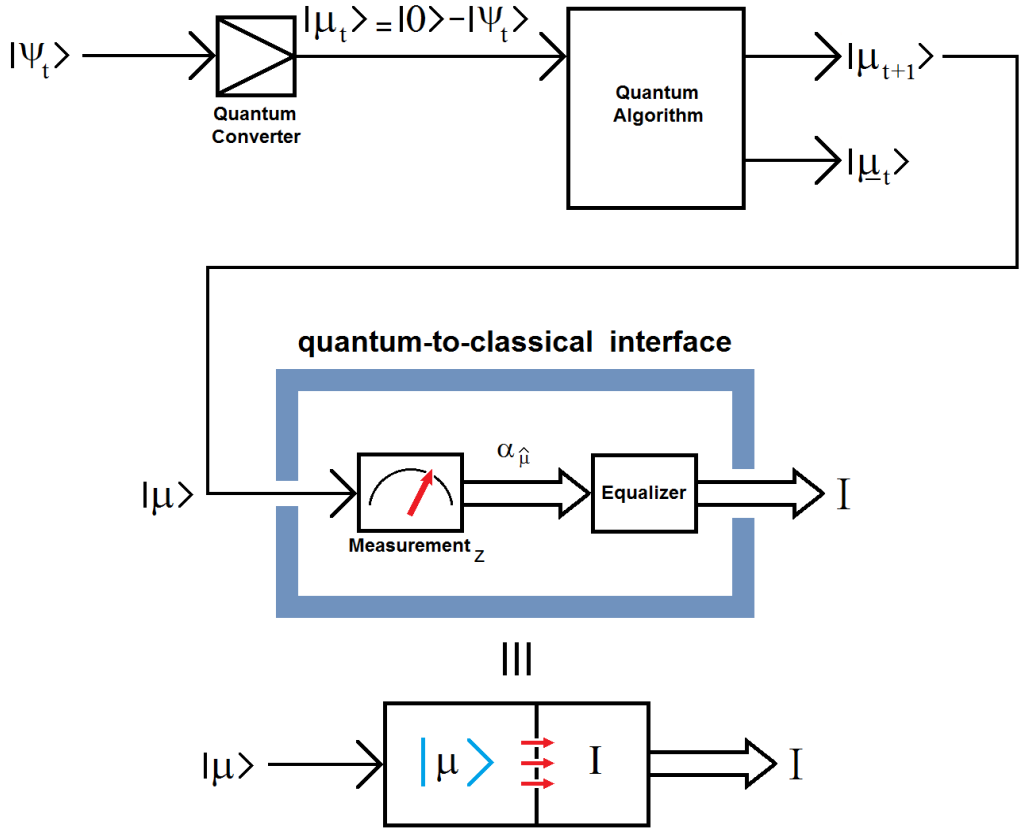

**Fig. 13** Quantum-to-classical interface (Mode II).

## 6.3 Classical-to-quantum interface

This interface is automatic and direct because we need the following correspondences, i.e.: $0 \to |0\rangle$ and $1 \to |1\rangle$, only. According to PAO, it is only necessary to measure the $z$ axis in Fig.14, as in the case of the above interfaces.

As we can see in such figure, this interface is essentially an automatic control system, with a set-point (for feedback) whose output goes to the equalizer. Hence the signal flow goes to the actuator and controller, and finally, to the Input Qubit Source (IQS), that is to say, the qubits factory. To the IQS output, we make a measurement by a quantum-to-classical interface. When the level of vertical component ($z$ axis) of qubits matches the corresponding value of the image (i.e., steady-state is reached) then the switch will close, and the qubits will pass to the input of an eventual quantum algorithm (circuit or gate). An important point to note is that this interface requires prior to its construction. See Fig.14.

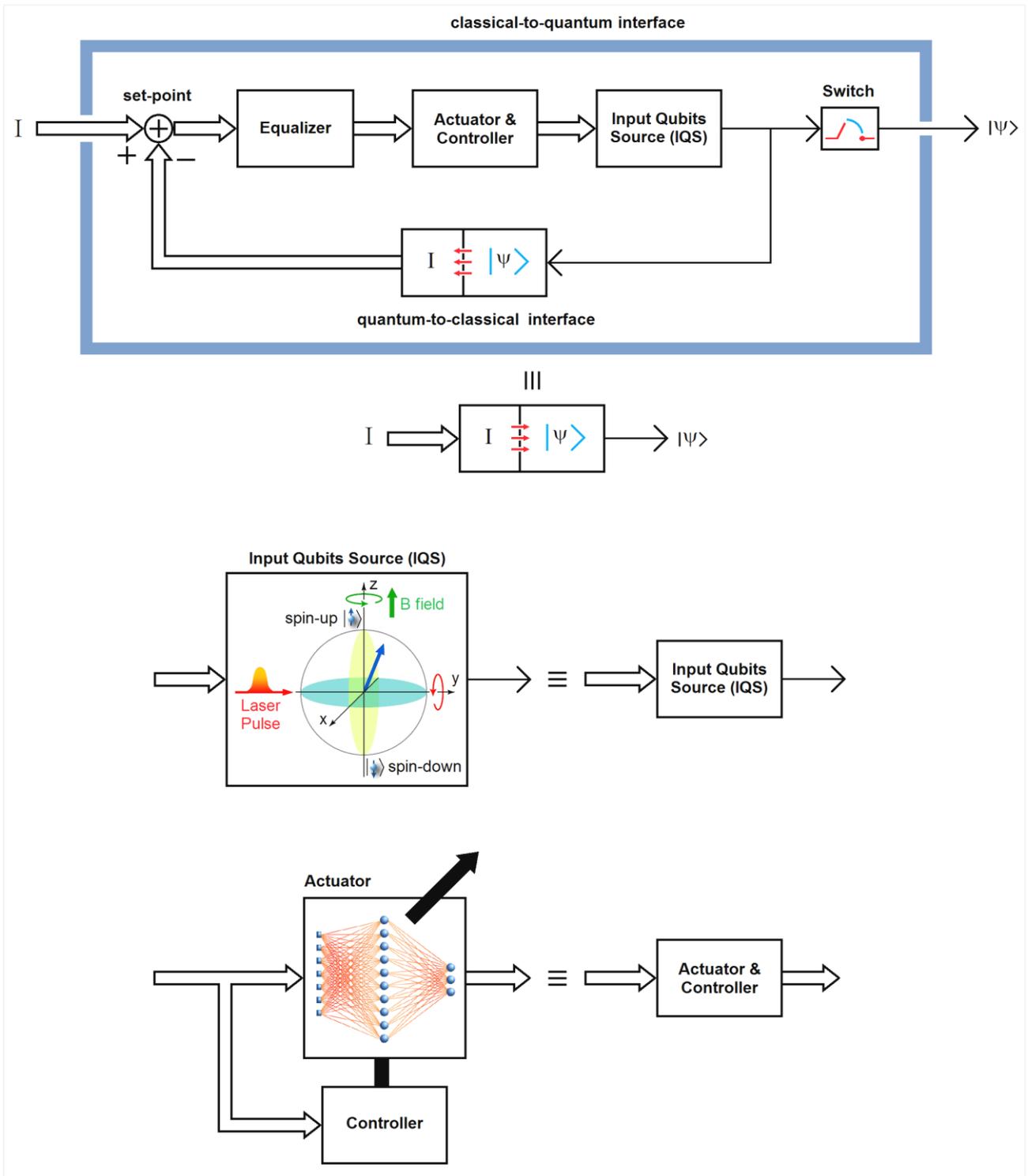

**Fig. 14** Classical-to-quantum interface.

## 7 New classical (Boolean) and quantum edge detectors

In this section, we present a method -exclusively- for Boolean and Quantum Boolean Image Processing, it is edge detection on Boolean (bitplane 7 or MSB) or quantum Boolean image. We present here it classical versions alone. Considering the above features of quantum logic and PAO criterion, the extension to the quantum version of this method is automatic. However, some preliminary considerations are necessary to understand the proceeding based on a convolutive mask.

## 7.1 Convolutive mask (Boolean or quantum)

In both cases (i.e., Boolean and quantum) we use an algorithm based on a convolutive masks with a horizontal rafter (see Fig. 15) on that image to which we must make an edge detection [65-67].

The main idea is to make an interaction between the mask and a portion of the image to be processed (with the same dimension as the mask) and that the result of said interaction to replace central pixel value of the image portion affected by the mask [14-17].

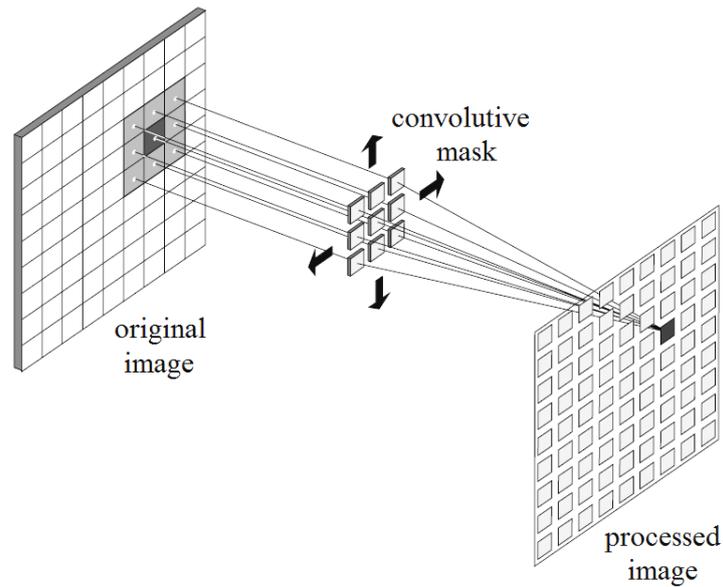

**Fig. 15** The convolution between the mask and the original image in a horizontal rafter produce the processed image.

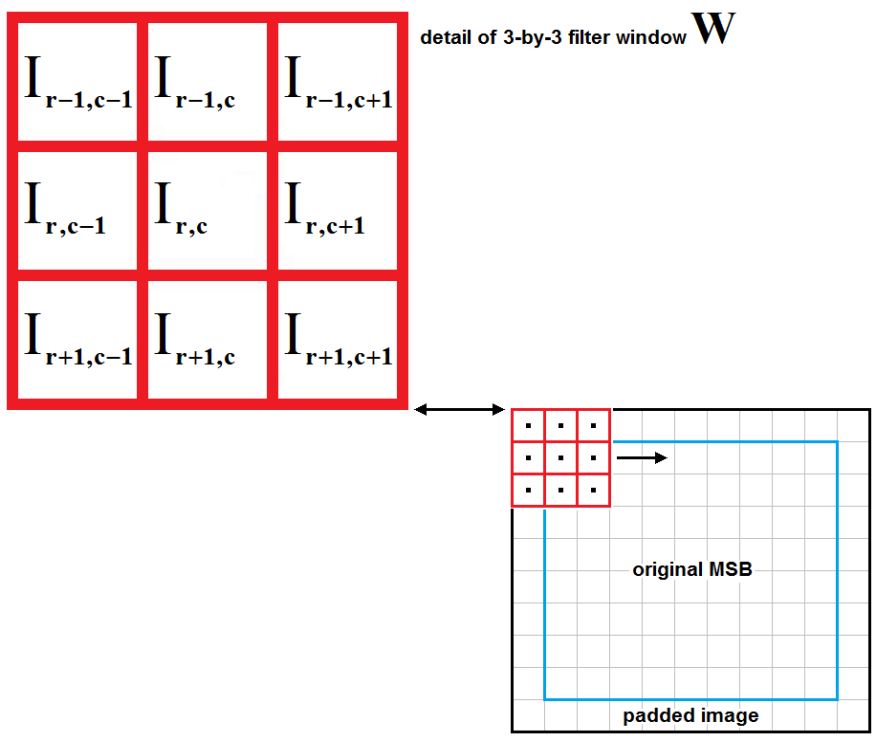

**Fig. 16** An example of 3x3 filter window for new edge-detection algorithm on a Image.

## 7.2 Classical (Boolean) New Edge Detector

Based on Fig.16, we take a mask of 3×3 (often called *kernel,* which should be of any size, that is, not only 3×3, provided it has the same number of rows and columns and the dimension is an odd number) which is applied in a horizontal rafter way.

This algorithm involves three steps based on Fig.16, namely:
1. Let's make OR Boolean operation between all elements of the kernel, except its center, i.e., I(r,c)
2. Let's make XOR Boolean operation between the previous step results and I(r,c)
3. Let's replace original I(r,c) with the results of the previous step results

Finally, we present the classical version (Boolean) in MATLAB® code of the new edge detection algorithm:

```
function I2 = bed(I)

% P = Number of times that it is applied
% w = Mask for rafter (window or kernel)
[ROW,COL] = size(I);
w = input('w = ');
P = input('p = ');
ROW2 = ROW+w-1;
COL2 = COL+w-1;
Iaux = zeros(ROW2,COL2);
Iaux(1+floor(w/2):ROW2-floor(w/2),1+floor(w/2):COL2-floor(w/2)) = I; % flaps for padded image
clear I;
Iaux2 = [];
for p = 1:P
 for r = 1+floor(w/2):ROW2-floor(w/2)
  for c = 1+floor(w/2):COL2-floor(w/2)
   for rw = 1:w % row kernel
    for cw = 1:w % column kernel
     W(rw,cw) = Iaux(r-(1+floor(w/2))+rw,c-(1+floor(w/2))+cw); % raw and column kernel
    end
   end
   aux = 0;
   for rw = 1:w % row kernel
    for cw = 1:w % column kernel
     if rw ~= 1+floor(w/2) & cw ~= 1+floor(w/2),
      aux = aux | W(rw,cw);
     end
    end
   end
   Iaux2(r,c) = xor(W(1+floor(w/2),1+floor(w/2)),aux);
  end
 end
end
clear Iaux;
I2 = Iaux2(1+floor(w/2):ROW2-floor(w/2),1+floor(w/2):COL2-floor(w/2));

return;
```

## 7.3 Quantum New Edge Detector

Based on PAO criterion and its logic, we are going to apply the same algorithm of last subsection, however, regarding to μ instead of α, i.e., there is a direct relationship between classical Boolean logic and quantum logic based on μ, and thanks to PAO criterion [49].

## 8 Metrics and Simulations

In in this section, we present a new metric especially designed for experiments which are developed here, and which consists in the comparison of classical and quantum version of denoising/despeckling and edge detection algorithms, outside and inside quantum computer, respectively.

### 8.1 Metric: Only One is Enough (OOIE)

This is a conspicuous metric for these cases, which it is a quantity used to measure how close forecasts or predictions are to eventual outcomes. The Only One is Enough (OOIE) for Boolean images (MSB) is given by

$$\text{OOIE} = \bigvee_{r,c} \left( I_{classical}(r,c) \wedge I_{quantum}(r,c) \right)$$
$$\forall r \in [1,R] \wedge \forall c \in [1,C], \text{ with } I_{classical} \in [0,1]^{R \times C} \text{ and } I_{quantum} \in [0,1]^{R \times C} \tag{23}$$

which for two $R \times C$ (rows-by-columns) images $I_{classical}$ and $I_{quantum}$, $I_{classical}$ means classical processed image, and $I_{quantum}$ means quantum processed image.

Typical values for the OOIE are 0 (zero) if both images are exactly equals and 1 (one) so only one element of them is different (i.e., only one is enough).

### 8.2 Simulations: Edge-detection of multimedia images

Edge detection (Fig. 17) where I is the original classical image (Boolean MSB), $|\psi\rangle$ is the quantum image, $|\psi_s\rangle$ is the quantum edge detected image, $I_{s,q}$ is the classical edge detected image from quantum edge-detection process, and $I_{s,c}$ is the classical edge detected image from a classical edge-detection process.

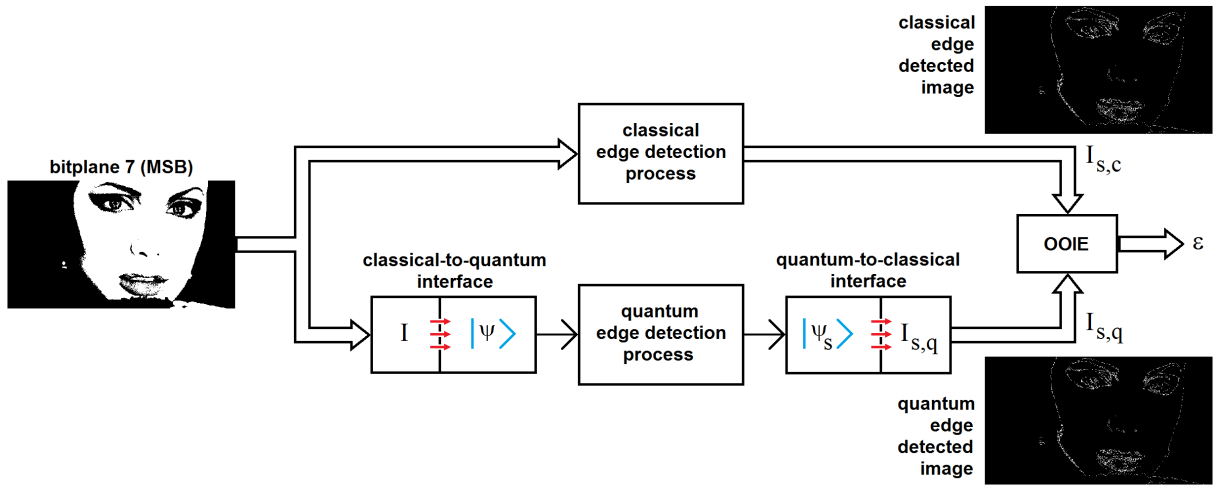

**Fig. 17** Edge detection for classical and quantum contexts.

For these experiments, all images are subjected to a built-in MATLAB® function *rgb2gray* explained before, that converts a color image into gray [50], and an own MATLAB® function *slicer* thanks to which we obtain all bitplanes, with special emphasis in the 7[th] one called MSB.

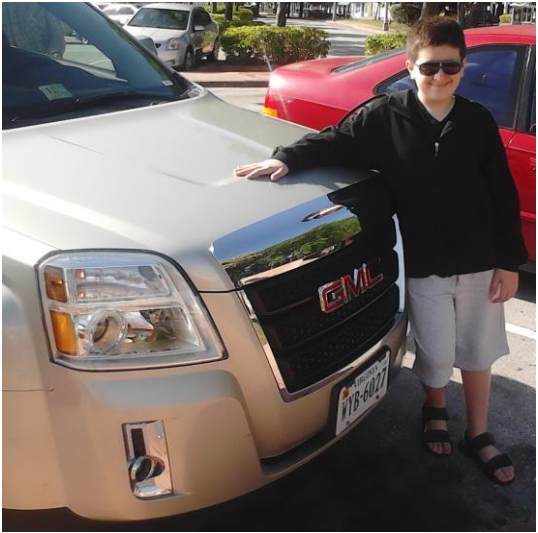
Original

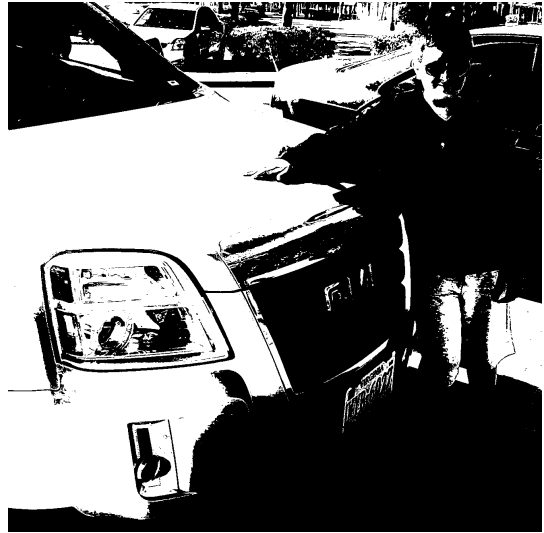
Bitplane 7 (MSB)

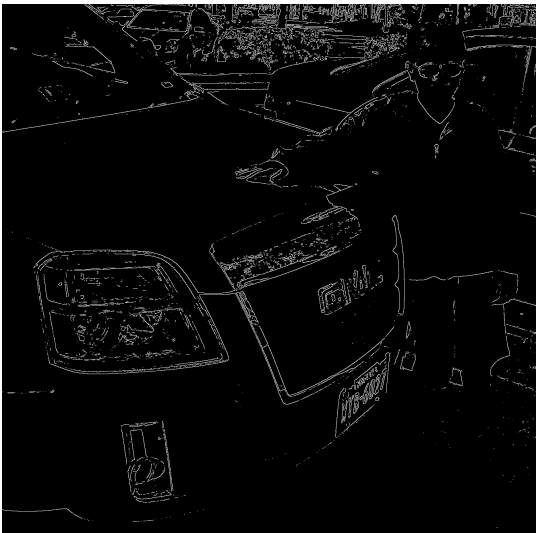
Classical Edge Detection

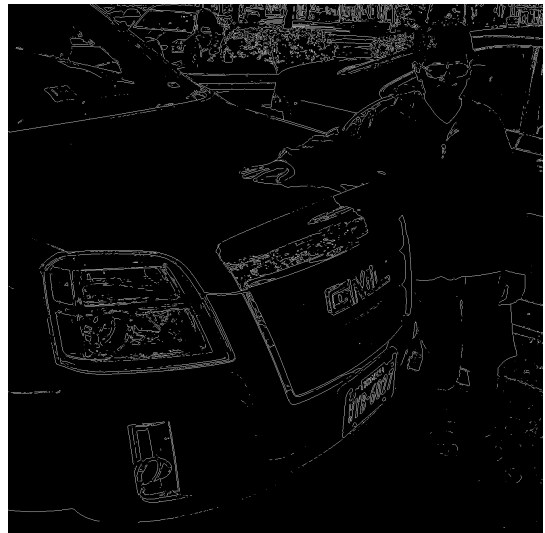
Quantum Edge Detection (μ)

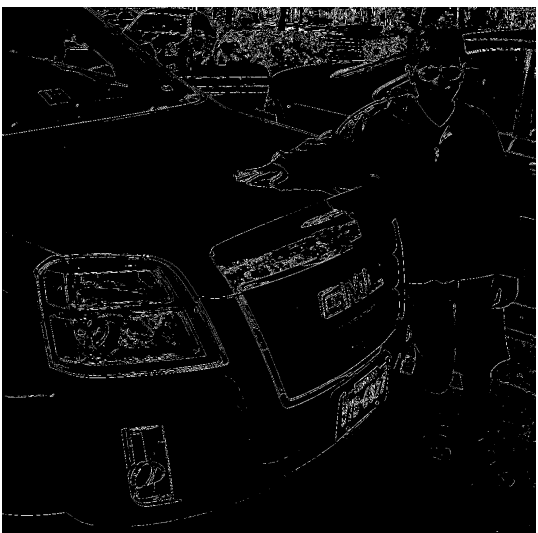
Error pixel-to-pixel

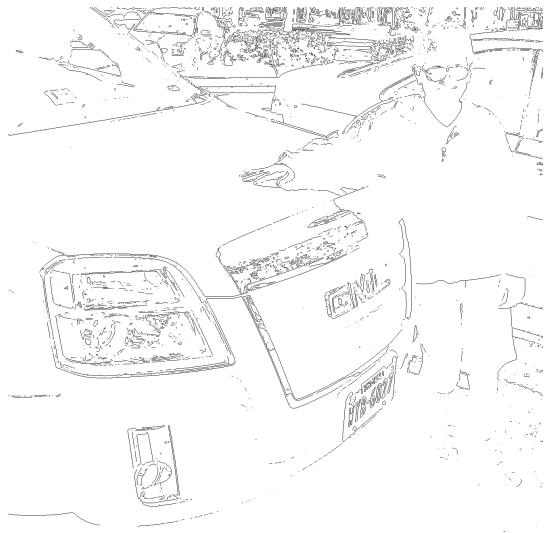
Quantum Edge Detection (α)

**Fig. 18** Edge Detection for Agus in Miami.

In Fig.18, we can see original image of *Agus in Miami* (top-left), bitplane 7 or MSB (top-right), classical version (Boolean) of edge detection (middle-left), quantum version of edge detection (middle-right) regarding to µ, Error pixel-to-pixel between classical and quantum processed images (down-left), and finally, quantum version of edge detection (down-right) regarding to α. Besides, the differences of Fig.18 (down-left) are significant.

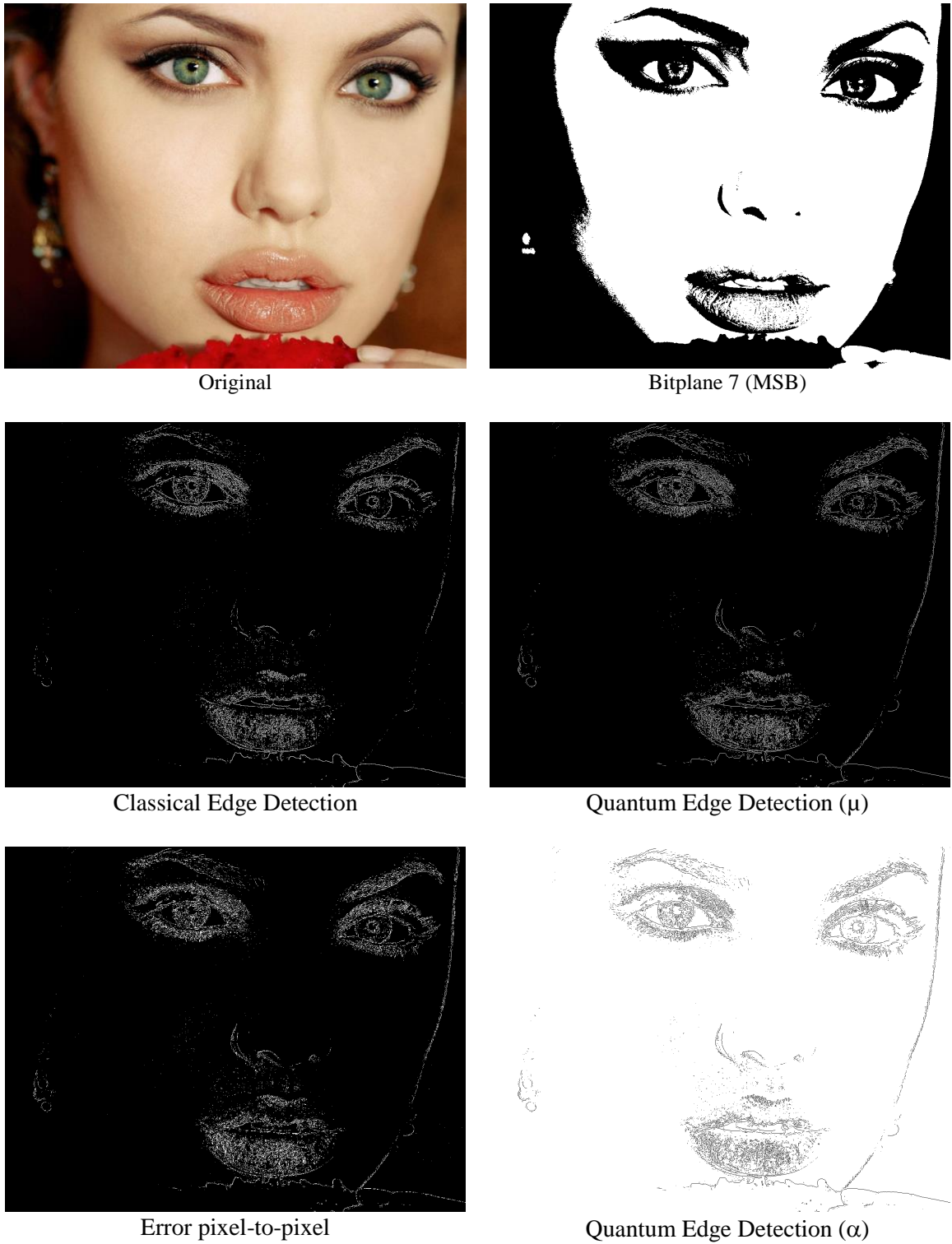

**Fig. 19** Edge Detection for Angelina.

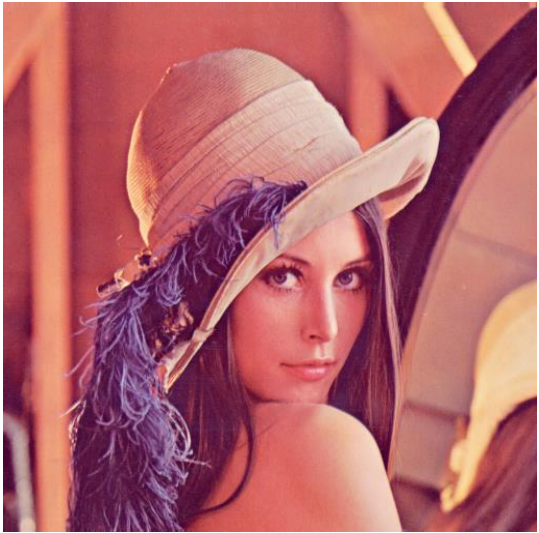
Original

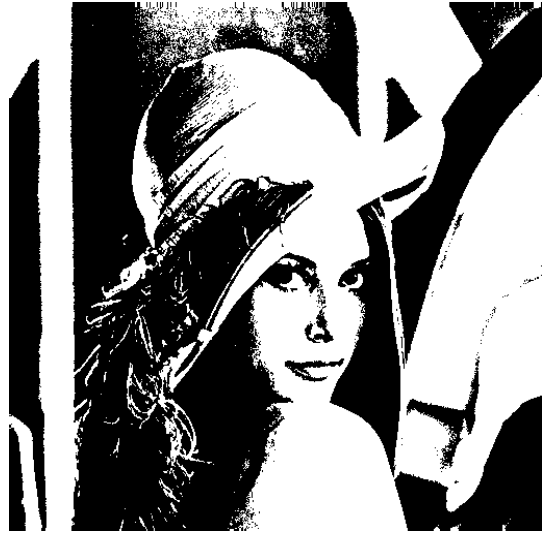
Bitplane 7 (MSB)

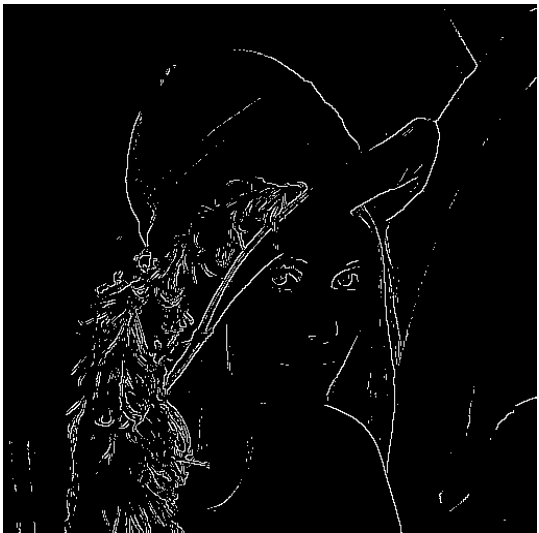
Classical Edge Detection

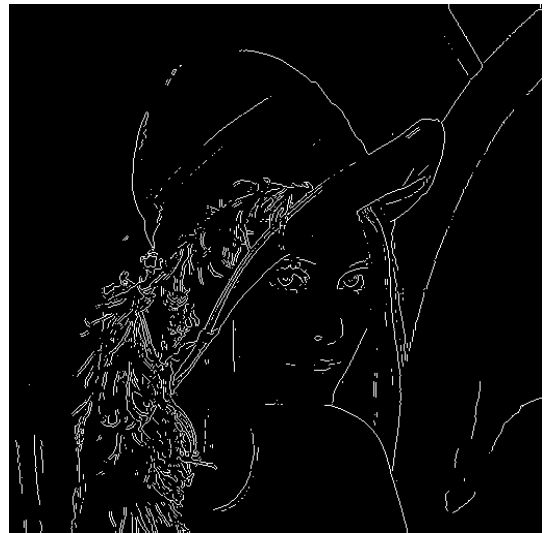
Quantum Edge Detection (μ)

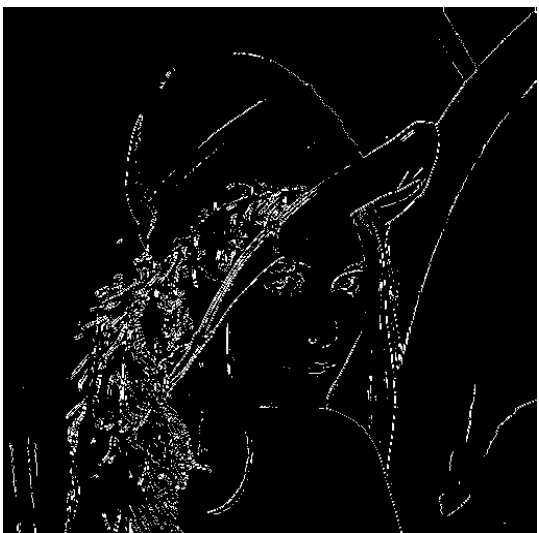
Error pixel-to-pixel

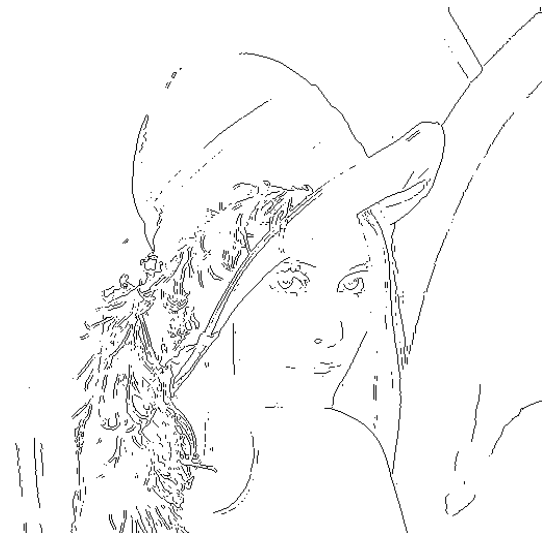
Quantum Edge Detection (α)

**Fig. 20** Edge Detection for Lena.

In Fig.19, we can see original image of *Angelina* (top-left), bitplane 7 or MSB (top-right), classical version (Boolean) of edge detection (middle-left), quantum version of edge detection (middle-right) regarding to μ, Error pixel-to-pixel between classical and quantum processed images (down-left), and finally, quantum version of edge detection (down-right) regarding to α. Besides, the differences of Fig.19 (down-left) are significant too like Fig.18.

Finally, we obtain similar results for *Lena*, however, and as it was predictable, *Lena* (Fig. 20) has the worst visual results of the three.

TABLE V
METRIC OF EDGE DETECTION: CLASSICAL VS. QUANTUM

| IMAGE | ONLY ONE IS ENOUGH (OOIE) |
|---|---|
| AGUS | 1 |
| ANGELINA | 1 |
| LENA | 1 |

Table V show us the metric for this experiment. Notice that OOIE are 1 (one) in all cases. However, in three cases the visual results are optimum. Unlike [49], where these differences were due to noise and state measurement, in this work these differences are due solely to implementation problems.

# 9 Conclusions and Future Works

In this paper we have presented -in order- the following advances:

9.1 Pole-to-pole Axis Only (PAO)

Basically, PAO consists of a new criterion and Logic based on projections onto vertical axis of Bloch's Sphere's exclusively. This approach allowed us:

1. a simpler development of logic quantum operations, where they will closer to those used in the classical digital image processing algorithms,
2. incorporation of quantum converter before quantum algorithm (circuit or gate), or classical converters after measurement of vertical projection for each qubit, which allow develop quantum logic operations (on a Hermitian context) as if it were in a classical context in the real field
3. building simple and robust classical-to-quantum and quantum-to-classical interfaces, based on the need to measure only the projections on the vertical axis (i.e., *z*-axis, or pole-to-pole axis)
4. build a simple and robust optimal estimator of quantum states avoiding Algebra Kroeneker

9.2 Quantum-Boolean Image Processing (QuBoIP)

This technique acts on Boolean and then quantum Boolean images, exclusively. To achieve this, we need to convert a color image to gray version. Then, we need get the bitplanes of the latter, extracting the MSB (bitplane 7) of the resulting set. This is followed by the classic-to-quantum interface, etc.

Summing-up, QuBoIP is a powerful tool to represent and process images within a quantum computer efficiently. Based on PAO criterion and considering that the measurement of a CBS does not alter the result, it is then presented as the only viable possibility of Quantum Image Processing implementation, in practice.

9.3 Classical-to-quantum and quantum-to-classical interfaces

We developed two modes of quantum-to-classical interface and one mode of classical-to-quantum interface according to the above (i.e., based on PAO). In addition to everything mentioned in the corresponding sec-tion, once obtained the $\mu$ based on its $\alpha$, to reach the levels of the external image (classical). Unlike [49], we don´t need an equalizer and a rounder if we work with bilevel images. These interfaces represent the purest functional interpretation on how they should work the same. This is the greatest contribution of PAO.

9.4 A new metric

In a special section on metric and simulations, a new metric based on the comparison between the classical and quantum versions algorithms for edge detection of images was presented. Notable differences between the results of classical and quantum versions of such algorithms (outside and inside of quantum computer, respectively) showed the need for a better implementation inside measurement scheme.

Summing-up, our quantum image processing for edge detection, however, it is far from the classical image processing as explained above. The later is one of the remaining tasks. The other is to apply the developed in this work to any quantum algorithm and Quantum Physics in general.

Finally, the classical technique (i.e., Boolean edge-detection) was implemented in MATLAB® R2014a (Mathworks, Natick, MA) [50] on a notebook with Intel® Core(TM) i5 CPU M 430 @ 2.27 GHz and 6 GB RAM on Microsoft® Windows 7© Home Premium 32 bits. Besides, a simulated version of quantum implementations were done on a GPU cluster, NVIDIA® Tesla© 2050 GPU [68] with a peak performance of approximately 500 GFLOPS, with an achieved performance of approximately 250 GFLOPS in OpenCL. The GPU needed approximately 2.5 GB of bandwidth with InfiniBand connectivity at quad data rate (QDR) QLogic® [69] or 40 Gb speeds.